\pdfoutput=1

\documentclass[11pt]{article}

\usepackage{EMNLP2022}
\usepackage{times}
\usepackage{latexsym}

\usepackage[T1]{fontenc}

\usepackage[utf8]{inputenc}

\usepackage{microtype}

\usepackage{enumitem}

\usepackage{xscommand}
\usepackage{MnSymbol}

\usepackage{subcaption}

\title{\name: Learning based Hybrid Local Search Algorithm for \\ Text  Hard-label Adversarial Attack}

\author{Zhen Yu\textsuperscript{1}\footnotemark[1] ,  \And Xiaosen Wang\textsuperscript{1,2}\thanks{\ \ The first two authors contributed equally.} ,  \And Wanxiang Che\textsuperscript{3}, \And Kun He\textsuperscript{1}\thanks{\ \ Corresponding author.}
  \AND
  \textnormal{\textsuperscript{1} School of Computer Science and Technology,}\\ Huazhong University of Science and Technology, Wuhan, China \\
  \textsuperscript{2} Huawei Singular Security Lab, Beijing, China\\
  \textsuperscript{3} Research Center for SCIR, Harbin Institute of Technology, Harbin, China \\
  {\small\texttt{\{baiding15,xiaosen\}@hust.edu.cn, car@ir.hit.edu.cn, brooklet60@hust.edu.cn} }} 

\begin{document}
\maketitle
\begin{abstract}

Existing textual adversarial attacks usually utilize the gradient or prediction confidence to generate adversarial examples, making it hard to be deployed in real-world applications. To this end, we consider a rarely investigated but more rigorous setting, namely hard-label attack, in which the attacker can only access the prediction label. 
In particular, we find  we can learn the importance of different words via the change on prediction label caused by word substitutions on the adversarial examples.
Based on this observation, we propose a novel adversarial attack, termed \textbf{Text} \textbf{Ha}rd-label atta\textbf{cker} (\textbf{\name}).
\name randomly perturbs lots of words to craft an adversarial example. Then, \name adopts a hybrid local search algorithm with the estimation of word importance from the attack history to 
minimize the adversarial perturbation.
Extensive evaluations for text classification and textual entailment show that \name significantly outperforms existing hard-label attacks regarding the attack performance as well as adversary quality. Code is available at \href{https://github.com/JHL-HUST/TextHacker}{https://github.com/JHL-HUST/TextHacker}.
\end{abstract}

\section{Introduction}
Despite the unprecedented success of Deep Neural Networks (DNNs), they are known to be vulnerable to adversarial examples~\cite{lbfgs}, in which imperceptible modification on the correctly classified samples could mislead the model. Adversarial examples bring critical security threats to the widely adopted deep learning based systems, attracting enormous attention on adversarial attacks and defenses in various domains, \eg Computer Vision (CV)
~\cite{lbfgs,fgsm,madry2018pgd,wang2021admix}
 and Natural Language Processing (NLP)~\cite{papernot2016crafting,threestrategies,pwws,Wang2022Randomized,Yang2022Robust}, \etc.

\par

Compared with adversarial attacks in CV,  textual adversarial attacks are more challenging due to the discrete input space and lexicality, semantics and fluency constraints. Recently, various textual adversarial attacks have been proposed, including white-box attacks~\cite{hotflip,textbugger,wang2021Adversarial}, score-based attacks~\cite{ga,sememepso} and hard-label attacks~\cite{textdecepter,hardlabelattack}. Among these methods, hard-label attacks that only obtain the prediction label are more realistic in real-world applications but also more challenging.


\par

Existing white-box attacks~\cite{textbugger,wang2021Adversarial} and score-based attacks~\cite{pwws,yang2020greedy} usually evaluate the word importance using either the gradient or change on logits after modifying the given word to craft adversarial examples. In contrast, due to the limited information (\ie only the prediction labels) for hard-label attacks, it is hard to estimate the word importance, leading to relatively low effectiveness and efficiency on existing hard-label attacks~\cite{hardlabelattack,ye2022texthoaxer}.

Zang \etal~\shortcite{rfhardlabel} have shown that estimating the word importance by reinforcement learning algorithm via the prediction confidence exhibits good attack performance for score-based attacks, but performs poorly for hard-label attacks. We speculate that it cannot effectively estimate the word importance via the prediction label since most of the times the label does not change when turning benign samples into adversaries. It inspires us to investigate 
the problem: \textit{How to effectively estimate the word importance using the prediction label?} In contrast, Wang \etal~\shortcite{wang2021Natural} show that replacing some words with synonyms could easily convert adversarial examples into benign samples. Thus, we could obtain abundant and useful information (\ie changes of prediction label) for word importance estimation by word substitutions on the adversarial examples during the attack process.
Such learned word importance could in turn guide us to minimize the word perturbation between adversarial examples and original samples. 


Based on the above observation, we propose a novel adversarial attack, named \textbf{Text} \textbf{Ha}rd-label atta\textbf{cker} (\textbf{\name}). \name contains two stages, namely adversary initialization and perturbation optimization.
At the adversary initialization stage, we substitute each word in the input text with its synonym iteratively till we find an adversarial example.
At the perturbation optimization stage, \name highlights the importance of each word based on the prediction label of the initialized adversarial example after synonym substitutions. Then \name adopts the hybrid local search algorithm with local search~\cite{aarts2003local} as well as recombination~\cite{radcliffe1993genetic} to optimize the adversarial perturbation using such word importance, and simultaneously updates the word importance based on the model output.

\par

To validate the effectiveness of the proposed method, we compare \name with two hard-label attacks~\cite{hardlabelattack, ye2022texthoaxer} and two evolutionary score-based attacks~\cite{ga,sememepso} for text classification and textual entailment. Empirical evaluations demonstrate that \name significantly outperforms the baselines under the same amount of queries, achieving higher average attack success rate with lower perturbation rate and generating higher-quality adversarial examples.

\section{Related Work}
This section briefly introduces the textual adversarial attacks and hybrid local search algorithm.

\subsection{Textual Adversarial Attacks}
Existing textual adversarial attacks fall into two settings: a) \textbf{white-box attacks}~\cite{threestrategies,textbugger,zhang2020generating,meng2020geometry,wang2021Adversarial} allow full access to the target model, \eg architecture, parameters, loss function, gradient, output, \etc. b) \textbf{black-box attacks} only allow access to the model output. Black-box attacks could be further split into two categories, in which \textbf{score-based attacks}~\cite{deepwordbug,ga,pwws,textfooler,rfhardlabel,sememepso,garg2020bae} could access the output logits (\ie prediction confidences) while \textbf{hard-label attacks}~\cite{textdecepter,hardlabelattack,ye2022texthoaxer} could only utilize the prediction labels.

Intuitively, hard-label attacks are much harder but more applicable in the real world and gain increasing interests. TextDecepter~\cite{textdecepter} hierarchically identifies the significant sentence among the input text and the critical word in the chosen sentence for attack. Hard label black-box attack (HLBB)~\cite{hardlabelattack} initializes an adversarial example via multiple random synonym substitutions and adopts a genetic algorithm to minimize the adversarial perturbation between the initialized adversarial example and original text. TextHoaxer~\cite{ye2022texthoaxer} randomly initializes an adversarial example  and optimizes the perturbation matrix in the continuous embedding space to maximize the semantic similarity and minimize the number of perturbed word between the current adversarial example and the original text.

Existing hard-label attacks access the prediction labels which are only used to evaluate adversarial examples without exploiting more information about the victim model. In this work, we learn the importance of each word \wrt the model based on the attack history, which is used to enhance the effectiveness of the attack.

\subsection{Hybrid Local Search Algorithm}

Hybrid local search algorithm is a popular population based framework, which is effective on typical combinatorial optimization problems~\cite{galinier1999hybrid}.
It usually contains two key components, \ie local search and recombination.
Given a population containing multiple initial solutions, the local search operator searches for a better one from the neighborhood of each solution to approach the local optima. The recombination operator crossovers the existing solutions to accept non-improved solutions so that it could jump out of the local optima. Then it adopts the fixed number of top solutions for the next iteration. Compared to other evolutionary algorithms, \eg genetic algorithm~\cite{anderson1994genetic},  particle swarm optimization~\cite{kennedy1995particle}, \etc, hybrid local search algorithm balances the local and global exploitation that helps explore the search space with much higher efficiency.

In this work, we follow the two-stage attack strategy in HLBB~\cite{hardlabelattack}. 
At the optimization stage, 
we utilize the word importance learned from the attack history to guide the local search and recombination.
Thus, our method can focus on more critical words in the neighborhood which helps us find the optimal adversarial example from the whole search space more efficiently.


\section{Methodology}
In this section, we first introduce the preliminary, symbols and definitions in \name, then provide a detailed description of  the proposed method. 
%


\subsection{Preliminary}
Given the input space $\mathcal{X}$ containing all the input texts and the output space $\mathcal{Y}=\{y_1, y_2, \ldots, y_k\}$, a text classifier $f\colon \mathcal{X} \to \mathcal{Y}$ predicts the label $f(x)$ for any input text $x = \langle w_1, w_2,\ldots,w_n \rangle \in \mathcal{X}$, in which $f(x)$ is expected to be equal to its ground-truth label $y_{true} \in \mathcal{Y}$. 
The adversary typically adds an imperceptible perturbation on the correctly classified input text $x$ to craft a textual adversarial example $x^{adv}$ that misleads classifier $ f$:
\begin{equation*}
f(x^{adv}) \ne f(x) = y_{true}, \quad \mathrm{s.t.} \quad d(x^{adv}, x) < \epsilon,
\label{eq:adv_object}
\end{equation*}
where $d(\cdot, \cdot)$ is a distance metric (\eg the $\ell_p$-norm distance or perturbation rate) that measures the distance between the benign sample and adversarial example, and $ \epsilon$ is a hyper-parameter for the maximum magnitude of perturbation. 
We adopt the perturbation rate as the distance metric:
\begin{equation*}
    d(x^{adv},x) = \frac{1}{n}\sum_{i=1}^n \mathbbm{1}(w_i^{adv} \ne w_i), 
\end{equation*}
where $\mathbbm{1}(\cdot)$ is the indicator function and $w_i \in x$, $w_i^{adv} \in x^{adv}$.
Given a correctly classified text $x$, we could reformulate the adversarial attack as minimizing the perturbation between benign sample and adversarial example while keeping adversarial:
\begin{equation}
    \argmin_{x^{adv}} d(x^{adv},x) \quad \mathrm{s.t.} \quad f(x^{adv}) \ne f(x). \label{eq:our_object}
\end{equation} 

In this work, we propose a novel hard-label attack, named \name, to craft textual adversarial examples by only accessing the prediction label $f(x)$ for any input sample $x$.  

\begin{figure*}[t]
\centering
\includegraphics[width=\textwidth]{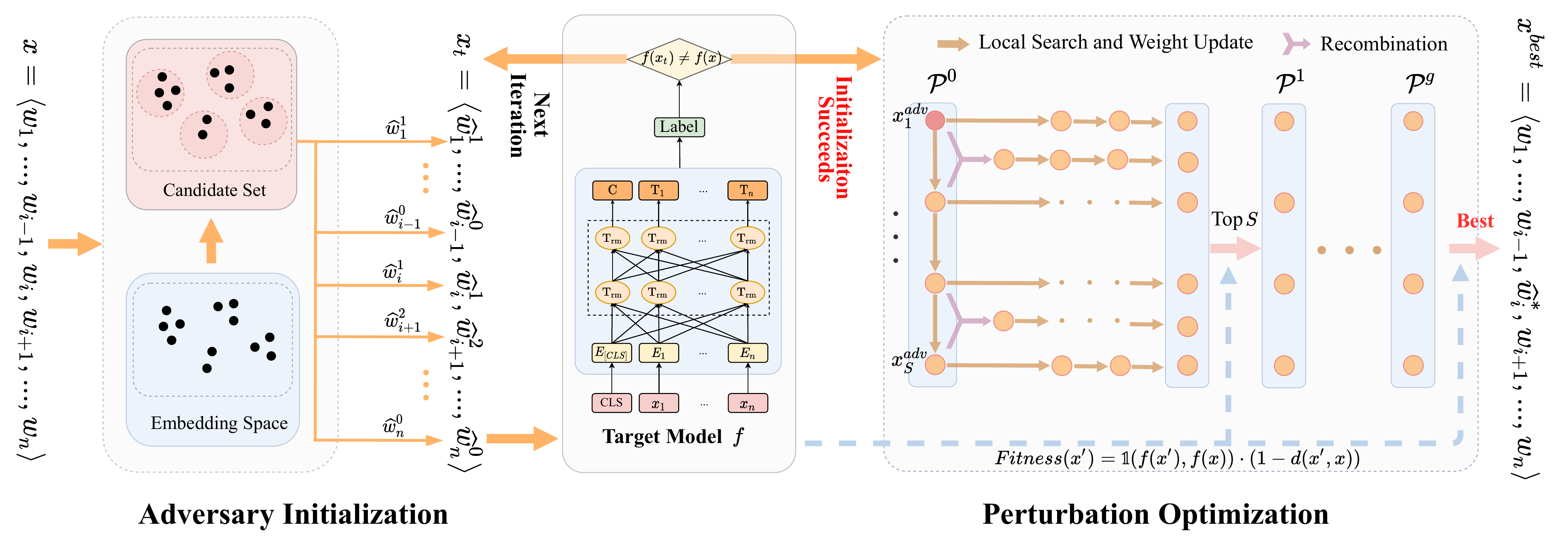}
\caption{The overall framework of the proposed \name algorithm.
\textbf{At the adversary initialization stage}, for a given input text $x$, after generating the candidate set for each word $w_i \in x$, we randomly substitute each word with its candidate words till we obtain an adversarial example $x^{adv}_1$. \textbf{At the perturbation optimization stage}, we first utilize local search to construct an initial population $\mathcal{P}^0$. Subsequently, we iteratively adopt recombination as well as local search to maximize the fitness function, and update the weight table after each local search.
}
\vspace{-0.3cm}
\label{fig:main_procedure}
\end{figure*}

\subsection{Symbols and Definitions}
\begin{itemize}[leftmargin=*]
    \item \textbf{Candidate set $\mathcal{C}(w_i)$}. For each word $w_i \in x$, we construct the candidate set $\mathcal{C}(w_i) = \{\hat{w}_i^0, \hat{w}_i^1,$ $\ldots,\hat{w}_i^m\}$ containing the word $w_i$ ($ \hat{w}_i^0 = w_i$) and its top $m$ nearest synonyms in the counter-fitted embedding space~\cite{mrkvsic2016counter}. All the substitutions would be constrained in this set.
    \item \textbf{Weight table $\mathcal{W}$}. We construct a weight table $\mathcal{W}$,  a matrix with the shape of $(n, m+1)$, in which each item $\mathcal{W}_{i,j}$ represents the word importance of $\hat{w}_i^j \in \mathcal{C}(w_i)$ and $\mathcal{W}_{i,:}=\sum_{j=0}^m \mathcal{W}_{i,j}$ denotes the position importance of word $w_i \in x$. The weight table $\mathcal{W}$ could guide the hybrid local search algorithm to determine the substitution at each iteration, which is initialized with all 0s.
    \item \textbf{$\delta$-neighborhood $N_\delta (x)$}. Given an input sample $x$, we define its $\delta$-neighborhood as the set of  texts in the input space $\mathcal{X}$ with at most $\delta$ different words from the sample $x$:
    \begin{equation*}
        N_{\delta}(x) = \{x^k \mid \sum_{i=1}^n \mathbbm{1}(w_i^k \ne w_i) \le \delta,\ x^k \in \mathcal{X}\},
    \end{equation*}
    where $w_i^k \in x^k, w_i \in x$ and $\delta$ is the maximum radius of the neighborhood. The neighborhood $N_{\delta}(x)$ reflects the search space
    for local search on input sample $x$.
    \item \textbf{Fitness function $F(x')$}. Given an input sample $x'$ and benign text $x$, we could define the fitness function as:
    \begin{equation}
    F(x') = \mathbbm{1}(f(x') \ne f(x)) \cdot (1-d(x',x)). \label{eq:fitness}
    \end{equation} 
    The fitness function could evaluate the quality of adversarial example to construct the next generation for \name. 
\end{itemize}

\subsection{The Proposed \name Algorithm}
As illustrated in Figure~\ref{fig:main_procedure}, \name contains two stages, \ie adversary initialization to initialize an adversarial example and perturbation optimization to minimize the adversarial perturbation.
In general, there are four operators used in \name, namely \operator{WordSubstitution} for adversary initialization,  \operator{LocalSearch}, \operator{WeightUpdate} and \operator{Recombination} for the hybrid local search algorithm at the perturbation optimization stage. The details of these operators are summarized as follows: 

\begin{itemize}[leftmargin=*]
\item \operator{WordSubstitution}$(x_t, \mathcal{C})$: Given an input text $x_t$ at $t$-th iteration with the candidate set $\mathcal{C}$ of each word $w_i \in x_t$, we randomly substitute each word $w_i \in x_t$ with a candidate word $\hat{w}_{i}^{j} \in \mathcal{C}(w_i)$ to craft a new text $x_{t+1}$. \operator{WordSubstitution} aims to search for an adversarial example in the entire search space by random word substitutions.

\item \operator{LocalSearch}$(x_t^{adv}, \mathcal{C}, \mathcal{W})$: As shown in Figure~\ref{fig:local_search}, for an adversarial example $x_t^{adv}$ at $t$-th iteration with the candidate set $\mathcal{C}$ and weight table $\mathcal{W}$, we randomly sample several (at most $\delta$) less important words $\hat{w}_i^{j_t} \in x^{adv}_t$ with the probability $p_i$ from all the perturbed words in $x_t^{adv}$:
\begin{equation*}
p_{i} = \frac{1 - \sigma(\mathcal{W}_{i,:})}{\sum_{i=1}^n \left[1- \sigma(\mathcal{W}_{i,:})\right]},
\end{equation*}
where $\sigma(x) = 1/(1+e^{-x})$ is the sigmoid function. 
The coarse-grained learning strategies in \operator{WeightUpdate} could easily make the gap between the word importance too large, resulting in probability distortion and getting stuck during the candidate word selection. To solve this problem, we utilize the sigmoid function with the saturation characteristic to reduce the excessive gap and make the probability more reasonable.
Then, we substitute each chosen word $\hat{w}_i^{j_t}$ with the original word $\hat{w}_i^0$ or with an arbitrary word $\hat{w}_i^{j_{t+1}} \in \mathcal{C}(w_i)$ using the probability $p_{i,j_{t+1}}$ equally to generate a new sample $x^{adv}_{t+1}$:
\begin{equation*}
p_{i,j_{t+1}} = \frac{\sigma(\mathcal{W}_{i,j_{t+1}})}{\sum_{j_{t+1}=0}^{m} \sigma(\mathcal{W}_{i,j_{t+1}})}.
\label{eq:word_prob}
\end{equation*} 

We accept $x_{t+1}^{adv}$ if it is still adversarial, otherwise we return the input adversarial example $x_{t}^{adv}$. 
\operator{LocalSearch} greedily substitutes unimportant word with the original word or critical word using the weight table to search for better adversarial example from the $\delta$-neighborhood of $x_t^{adv}$.

\begin{figure}[tb]
\centering
\includegraphics[width=0.43\textwidth]{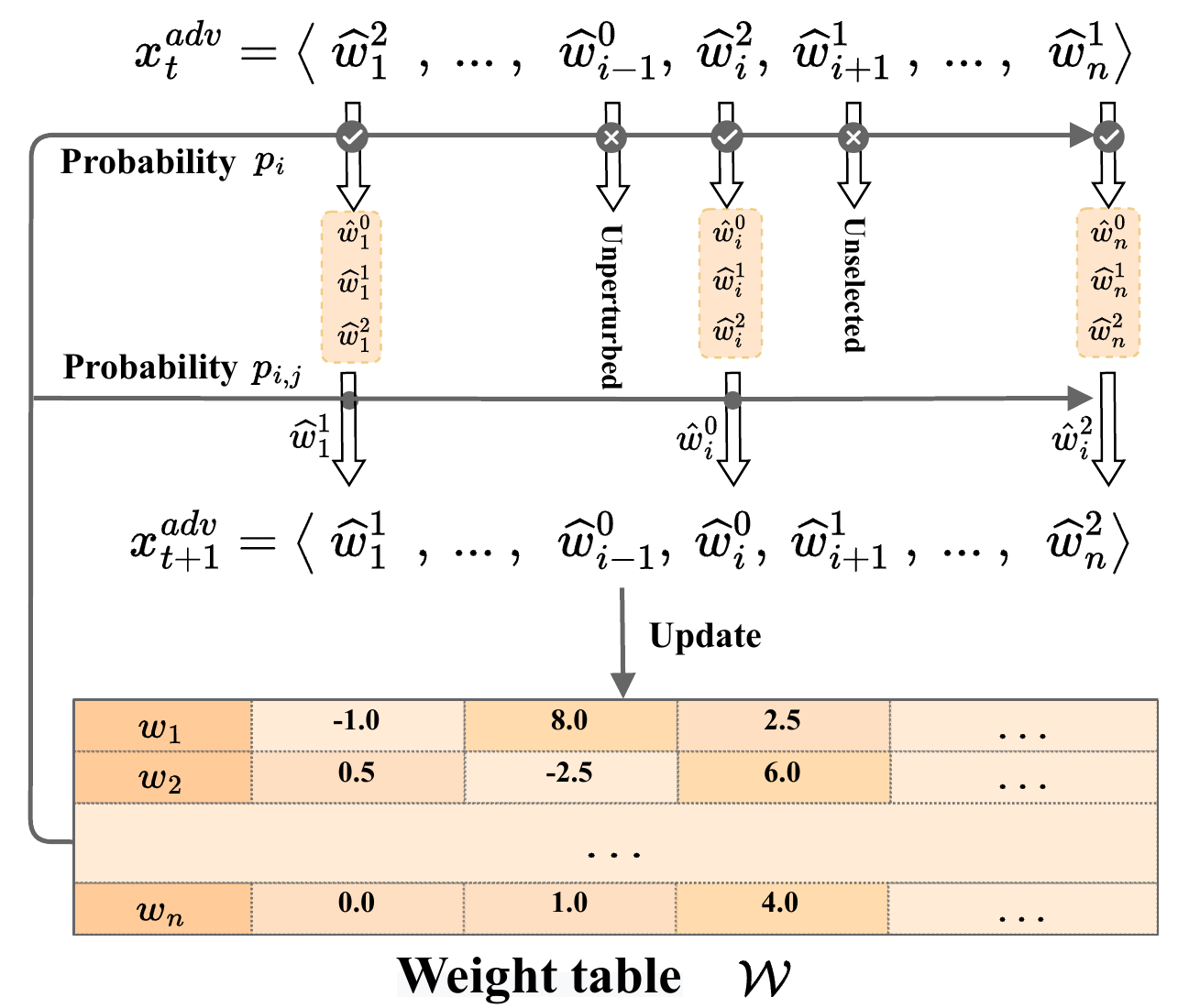}
\caption{The overview of the \operator{LocalSearch} and \operator{WeightUpdate}. For an adversary $x_t^{adv}$, we sample several words with  probability $p_i$ based on the weight table. Then, we substitute each sampled word with original word or its candidate word with  probability $p_{i,j}$ to generate a new text $x_{t+1}^{adv}$. Finally, we use the prediction label of the new text $x_{t+1}^{adv}$ to update the weight table.}
\label{fig:local_search}
\end{figure}

\item \operator{WeightUpdate}$(x_t^{adv}, x_{t+1}^{adv}, f, \mathcal{W})$: Given an adversarial example $x_t^{adv}$ at $t$-th iteration with the generated adversary $x_{t+1}^{adv}$ by local search, we update the word importance of each operated word $\hat{w}_i^{j_t} \in x_t^{adv}$ and $\hat{w}_i^{j_{t+1}} \in x_{t+1}^{adv}$, and the position importance of $w_i$ using the following rules:

\Rule{1}: For each replaced word $\hat{w}_i^{j_{t+1}}$, if $x_{t+1}^{adv}$ is still adversarial, it has positive impact on the adversary generation. So we increase its weight $\mathcal{W}_{i,j_{t+1}}$, and vice versa.

\Rule{2}: For each operated position $i$, if $x_{t+1}^{adv}$ is still adversarial, it has little impact on the adversary generation. So we decrease the position weight $\mathcal{W}_{i,:}$, and vice versa.

Specifically, if $x_{t+1}^{adv}$ is still adversarial, we assign the positive reward $r$ to each replaced word $\hat{w}_i^{j_{t+1}}$ using \Rule{1}, and reward $-2r$ to each $\hat{w}_i^{j_t}$ to decrease the weight summation  $\mathcal{W}_{i,:} = \sum_{j=0}^m \mathcal{W}_{i,j}$ in each operated position $i$ using \Rule{2}:
\begin{equation*}
    \begin{split}
        \mathcal{W}'_{i,j_{t+1}} &= \mathcal{W}_{i,j_{t+1}} + r, \mathrm{~~~~ } \mathcal{W}'_{i,j_{t}} = \mathcal{W}_{i,j_{t}} - 2r, \\
      \end{split}
\end{equation*}
where $r$ is the predefined reward value and $\mathcal{W}'$ is the weight table after this update.
Otherwise, we assign the reward $-r$ to each $\hat{w}_{i}^{j_{t+1}}$ and $2r$ to each $\hat{w}_{i}^{j_t}$.
\operator{WeightUpdate} highlights the important words and positions by assigning different reward for each operated word, which helps the \operator{LocalSearch} select more critical positions and synonyms to substitute.

\item \operator{Recombination}$(\mathcal{P}^t, \mathcal{W})$:  For 
the $t$-th generation population $\mathcal{P}^t$ that contains multiple adversarial examples, we combine two randomly sampled texts $x^a=\langle w_1^a, w_2^a, \ldots, w_n^a\rangle \in \mathcal{P}^t$ and $x^b=\langle w_1^b, w_2^b, \ldots, w_n^b\rangle \in \mathcal{P}^t$ to construct a recombined text $x^c=\langle w_1^c, w_2^c,\ldots, w_n^c\rangle$, where each word $w_i^c$ is randomly sampled from $\{w_i^a, w_i^b\}$ based on their weights in the weight table $\mathcal{W}$. We repeat the operation $|\mathcal{P}^t|/2$ times, and then return all the recombined texts.
\operator{Recombination} crafts non-improved solutions by randomly mixing two adversarial examples, which globally changes the text to avoid poor local optima.
\end{itemize}

In summary, as shown in Figure~\ref{fig:main_procedure}, at the adversary initialization stage, for an input text $x$, we adopt \operator{WordSubstituion} iteratively to search for an adversarial example. At the perturbation optimization stage, we initialize the weight table $\mathcal{W}$ and adopt the hybrid local search algorithm to minimize the adversary perturbation. Specifically, we first utilize the \operator{LocalSearch} to construct an initial population. At each iteration, we adopt \operator{Recombination} and \operator{LocalSearch} to generate several adversarial examples using the weight table $\mathcal{W}$. Then we utilize the fitness function in Equation~\eqref{eq:fitness} to filter adversarial examples for the next generation. After the adversary optimization, the adversary with the highest fitness would be regarded as the final adversarial example. The overall algorithm of \name is summarized in Algorithm~\ref{alg:main_alogithm}.

\algnewcommand{\LineComment}[1]{ \(\triangleright\) #1}
\begin{algorithm}[t] 
    \small
    \DontPrintSemicolon
    \newcommand\mycommfont[1]{\footnotesize\ttfamily{#1}}
    \SetCommentSty{mycommfont}
    \LinesNumbered
    \caption{The \name Algorithm}
	\label{alg:main_alogithm}
	\KwIn{Input sample $x$, target classifier $f$,  query budget  $T$, reward $r$, population size $S$, maximum number of local search $N$}
	\KwOut{Attack result and adversarial example}  
	\LineComment{\textbf{Adversary Initialization}}\\
	Construct the candidate set $\mathcal{C}(w_i)$ for each $w_i \in x$  \\
	
	$x_1 = x$, $x_1^{adv}=\mathrm{None}$ \\
	\For{$t=1 \to T$}{
	    $x_{t+1} = $ \operator{WordSubstituion}$(x_{t}, \mathcal{C})$  \\
	    \If{$f(x_{t+1}) \neq f(x)$}{
	       $x_1^{adv}=x_{t+1}$;
	        break   \\
	    }
	}
	\If{$x_1^{adv}$ is $\mathrm{None}$}{
	    \Return False, $\mathrm{None}$  \algorithmiccomment{Initialization fails} 
	}
	\LineComment{\textbf{Perturbation Optimization}}\\
    Initialize the weight table $\mathcal{W}$ with all 0s\\
    $x_{i+1}^{adv}=$\operator{LocalSearch}$(x^{adv}_i, \mathcal{C}, \mathcal{W})$ \\
    $\mathcal{P}^1=\{x_1^{adv}, \cdots, x_i^{adv}, \cdots,  x_S^{adv}\}$ \\
    $\ t = t + S - 1$;$\ g = 1$  \\
    \While{$t \le T$}{
        $\mathcal{P}^g=\mathcal{P}^g \cup \{\text{\operator{Recombination}}(\mathcal{P}^g, \mathcal{W})\}$ \\
	    \For{each text $x^{adv}_g \in \mathcal{P}^g$ }{
	        With $x^{adv}_{1} = x^{adv}_g$ for $i=1 \to N$: \\
	        \quad $x^{adv}_{i+1} = $ \operator{LocalSearch}$(x^{adv}_{i}, \mathcal{C}, \mathcal{W})$; \\ 
	        \quad \operator{WeightUpdate}$(x^{adv}_{i},x^{adv}_{i+1}, f, \mathcal{W})$ \\
	        $\mathcal{P}^g = \mathcal{P}^g \cup \{x_{N+1}^{adv}\}$ \\
	        $t = t +N$
	    }
	   Construct $\mathcal{P}^{g+1}$ with the top $S$ fitness in $\mathcal{P}^g$ based on Equation~\eqref{eq:fitness}\\
	    Record global optima $x^{best}$ with the highest fitness\\
	    $g = g + 1$ \\
	}
    \Return True, $x^{best}$ \algorithmiccomment{Attack succeeds}
\end{algorithm}

\section{Experiments}
In this section, we conduct extensive experiments on eight benchmark datasets and four models to validate the effectiveness of \name.



\begin{table*}[tb]
\centering
\small
\resizebox{\linewidth}{!}{
\begin{tabular}{clcccccccccccccc} 
\toprule
\multirow{2.4}*{\textbf{Model}} & \multirow{2.4}*{\textbf{Attack}} & \multicolumn{2}{c}{\textbf{AG's News}} &  &
\multicolumn{2}{c}{\textbf{IMDB}} & &
\multicolumn{2}{c}{\textbf{MR}} &  &
\multicolumn{2}{c}{\textbf{Yelp}} &  &
\multicolumn{2}{c}{\textbf{Yahoo! Answers}} \\

\cmidrule(lr){3-4}\cmidrule(lr){6-7}\cmidrule(lr){9-10}\cmidrule(lr){12-13}\cmidrule(lr){15-16}

~ & ~   & \textbf{Succ.} & \textbf{Pert.}     && \textbf{Succ.} & \textbf{Pert.}      && \textbf{Succ.} & \textbf{Pert.}   && \textbf{Succ.} & \textbf{Pert.}     && \textbf{Succ.} & \textbf{Pert.} \\ 

\midrule
\multirow{4}*{\textbf{BERT}}& GA & 40.5 & 13.4 &&  50.9 & ~~5.0 && 65.6  & 10.9 && 36.6  & ~~8.6 &&  64.2 & ~~7.6 \\ 
~  & PSO &  45.8  & 12.1 && 60.3  & ~~3.7    &&  74.4  & 10.7    && 47.9  & ~~7.5   && 64.7  & ~~6.6 \\
\cdashline{2-16}[2pt/2pt]
\specialrule{0em}{1.5pt}{1.5pt}
~  & HLBB  & 54.7  & 13.4 && 77.0  & ~~4.8   && 65.8  & 11.4   && 57.1   & ~~8.2    &&  82.0  & ~~7.7 \\
~  & TextHoaxer  & 52.0  & 12.8 && 78.8  & ~~5.1   && 67.1  & \textbf{11.1}   && 58.3   & ~~8.5    &&  83.1  & ~~7.6 \\
~ & \textbf{\name}   & \textbf{63.2}  & \textbf{11.9}   && \textbf{81.5} & ~~\textbf{3.4} && \textbf{73.1}  & 11.4   && \textbf{63.2} & ~~\textbf{6.7}  && \textbf{87.2}  & ~~\textbf{6.3} \\

\midrule
\multirow{4}*{\shortstack[c]{\textbf{Word}\\ \textbf{CNN}}}& GA & 70.0  & 12.1    && 59.6 & ~~5.9   & & 72.9  & 11.1     && 44.4  &  ~~9.0     && 62.0  & ~~8.7  \\
~  & PSO  & 83.5 & 10.4   && 55.6 & ~~4.2    && 80.7 & 10.7   && 45.6 & ~~7.4    && 52.7 & ~~7.0 \\
\cdashline{2-16}[2pt/2pt]
\specialrule{0em}{1.5pt}{1.5pt}
~  & HLBB  & 74.0 & 11.7  && 74.0 & ~~4.2    && 71.1 & 11.2    && 67.1 & ~~7.6    &&  78.7 & ~~7.8 \\
~  & TextHoaxer  & 73.5  & 11.5 && 76.5  & ~~4.6   && 71.1  & \textbf{10.7}   && 68.1   & ~~8.0    &&  78.6  & ~~7.8 \\
~ & \textbf{\name} & \textbf{81.7} &  \textbf{10.2} && \textbf{77.8} &  ~~\textbf{3.0}  && \textbf{78.3} & 11.1   && \textbf{75.4} & ~~\textbf{6.4}   && \textbf{84.5} & ~~\textbf{6.3}  \\

\midrule
\multirow{4}*{\shortstack[c]{\textbf{Word}\\ \textbf{LSTM}}}& GA & 45.5  &  12.4 & &  50.8  & ~~5.7   && 67.2  & 11.2  && 40.7  & ~~8.1 && 51.2  & ~~8.6 \\
~  & PSO & 54.2 & 11.6 && 42.5  & ~~4.5 &&  73.0 & 10.9 && 44.5  & ~~6.7 && 43.3  & ~~7.3 \\
\cdashline{2-16}[2pt/2pt]
\specialrule{0em}{1.5pt}{1.5pt}
~  & HLBB  & 56.8 & 12.7 &&  72.1 & ~~4.1 && 68.3  & 11.2 &&  61.0 & ~~6.6 && 70.8  & ~~8.3 \\
~  & TextHoaxer  & 56.5  & 12.3 && 73.5  & ~~4.5   && 67.9  & \textbf{10.7}   && 61.8   & ~~6.7    &&  70.1  & ~~8.1 \\
~ & \textbf{\name} & \textbf{64.7}  & \textbf{11.2} && \textbf{76.2}  & ~~\textbf{3.0} && \textbf{75.2}  & 11.2 && \textbf{65.4}  & ~~\textbf{5.5} &&  \textbf{75.5} & ~~\textbf{6.9} \\

\bottomrule
\end{tabular}
}
\caption{Attack success rate (Succ., \%) $\uparrow$, perturbation rate (Pert., \%) $\downarrow$ of various attacks  on three models using five datasets for text classification under the query budget of 2,000. $\uparrow$ denotes the higher the better. $\downarrow$ denotes the lower the better. We \textbf{bold} the highest attack success rate and lowest perturbation rate among the hard-label attacks.}
\label{tab:class_performance}
\end{table*}

\subsection{Experimental Setup}
\par

\textbf{Datasets}. 
We adopt five widely  investigated datasets, \ie AG's News~\cite{ag_and_yahoo}, IMDB~\cite{imdb}, MR~\cite{mr}, Yelp~\cite{ag_and_yahoo}, and Yahoo! Answers~\cite{ag_and_yahoo} for text classification. For textual  entailment,
we select SNLI~\cite{snli} and MulitNLI~\cite{mnli}, where MulitNLI includes matched version (MNLI) and mismatched version (MNLIm).

\textbf{Baselines}. We take the hard-label attacks HLBB~\cite{hardlabelattack} and TextHoaxer~\cite{ye2022texthoaxer} as our baselines. Since there are only few hard-label attacks proposed recently, we also adopt two evolutionary score-based attacks, \ie GA~\cite{ga} and PSO~\cite{sememepso} for reference, which extra utilize the prediction confidence for attack.

\textbf{Victim Models}. We adopt WordCNN~\cite{wordcnn}, WordLSTM~\cite{wordlstm}, and BERT base-uncased~\cite{bert} models for text classification and BERT base-uncased model for textual entailment.

\textbf{Evaluation Settings}. For \name, we set the neighborhood size $\delta = 5$, reward $r = 0.5$, population size $S=4$, maximum number of local search $N=8$. The parameter studies are given in Appendix~\ref{appendix:parameter}. For a fair comparison, we adjust the population size and adopt the same values for other parameters
as in their original papers to achieve better performance for the score-based attacks of GA and PSO. All the evaluations are conducted on 1,000 randomly sampled texts from the corresponding testset. We set the synonym number $m=4$. The attack succeeds if the perturbation rate of the generated adversarial example is smaller than $25\%$ to ensure the semantic constraints of the  adversarial examples.
As the task complexity varies across datasets, we set different query budget $T$ (\ie the maximum query number to the victim model) for different tasks (2,000 for text classification and 500 for textual entailment). The results are averaged on five runs to eliminate randomness.

\subsection{Evaluation on Attack Effectiveness}

We first conduct evaluations for text classification using five datasets on three models under the same query budget of 2,000. The results, including attack success rate and perturbation rate, are summarized in Table~\ref{tab:class_performance}. 
We could observe that \name consistently achieves higher attack success rate with lower perturbation rate across almost all the datasets and victim models than the hard-label attacks. 
Even for the score-based attacks of GA and PSO, \name exhibits better attack performance on most datasets and victim models.

\begin{table}[tb]
\setlength\tabcolsep{4pt}
\centering
\small
\begin{tabular}{lcccccc} 
\toprule
\multirow{2.4}*{\textbf{Attack}} & \multicolumn{2}{c}{\textbf{SNLI}} &
\multicolumn{2}{c}{\textbf{MNLI}} &
\multicolumn{2}{c}{\textbf{MNLIm}}\\

\cmidrule(lr){2-3}\cmidrule(lr){4-5}\cmidrule(lr){6-7}

~   & \textbf{Succ.} & \textbf{Pert.}     & \textbf{Succ.} & \textbf{Pert.}      & \textbf{Succ.} & \textbf{Pert.}  \\ 

\midrule
GA  & 67.2 & 14.6 & 67.6 & 12.6 & 66.9 & 12.2\\ 
PSO & 70.7 & 15.0  & 72.0  & 12.9  & 70.8 & 12.4 \\
\cdashline{1-7}[2pt/2pt]
\specialrule{0em}{1.5pt}{1.5pt}
HLBB  & 57.2 & \textbf{14.0}  & 58.3 & \textbf{12.2}  & 58.6 & \textbf{11.8} \\
TextHoaxer  & 61.0 & 14.1  & 64.0 & 12.4  & 63.8 & 12.0 \\
\textbf{\name}  & \textbf{70.3} & 15.0  & \textbf{68.3} & 12.8 &  \textbf{69.0} & 12.4\\

\bottomrule
\end{tabular}
\caption{Attack success rate (Succ., \%) $\uparrow$, perturbation rate (Pert., \%) $\downarrow$ of \name and the baselines on BERT using three datasets for textual entailment under the query budget of 500.}
\vspace{-0.3cm}
\label{tb:entail_performance}
\end{table}







To further validate the effectiveness of the proposed \name, we also conduct evaluations on BERT for three textual entailment tasks. As shown in Table~\ref{tb:entail_performance}, under the same query budget of 500,
\name outperforms HLBB by a clear margin of 10.0\%-13.1\% and TextHoaxer by 4.3\%-9.3\% on three datasets with similar perturbation rate. Compared with the score-based attacks, \name achieves lower attack success rate than PSO, but still gains better attack success rate than GA. It is acceptable since GA and PSO extra utilize the changes on prediction confidence introduced by synonym substitution, making the attack much easier than the hard-label attacks.


In conclusion, under the same query budgets, the proposed \name exhibits much better attack performance than existing hard-label attacks, for either text classification or textual entailment, and achieves comparable or even better attack performance than the advanced score-based attacks.

\subsection{Evaluation on Attack Efficiency}
\begin{figure}[t]
\centering
\includegraphics[width=0.43\textwidth]{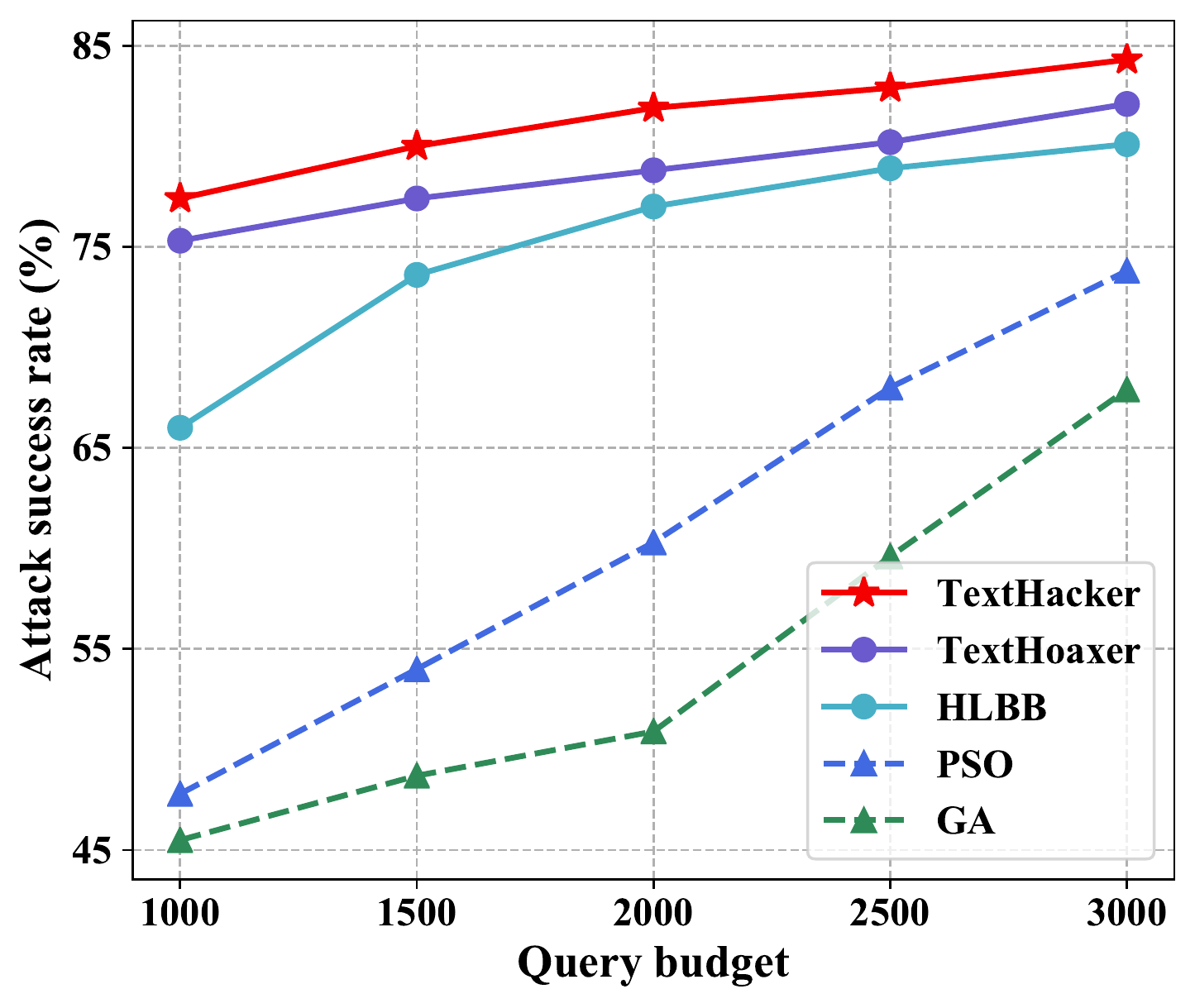}
\caption{Attack success rate (\%) $\uparrow$ of various attacks on BERT using IMDB dataset under various query budgets.
}
\vspace{-0.3cm}
\label{fig:query_number}
\end{figure}

In practice, the victim could block the attack by simply denying the access if they detect overload access within a short period. Hence, the attack efficiency, which often refers to the query budget for victim model, plays a key role in evaluating the effectiveness of black-box attacks.
On the other hand, the query budget significantly affects the attack performance of the algorithm. Thus, a good attack should exhibit consistent and superior attack performance under various query budgets.

We report the attack success rate of \name and the baselines under various query budgets on BERT using IMDB dataset in Figure~\ref{fig:query_number}. \name, HLBB and TextHoaxer exhibit remarkably higher attack success rate than GA and PSO under the limited query budget ($\leq 2,000$). We further analyze why GA and PSO perform poorly under the limited query budget in Appendix~\ref{sec:baseline_analysis}. When we continue to increase the query budget, the attack success rate of GA and PSO starts to increase rapidly but is still lower than that of \name, which maintains stable and effective performance. In general, \name consistently exhibits better attack performance under various query budgets, which further demonstrates the superiority of \name.

\begin{table}[tb]
\centering
\small
\begin{tabular}{lcccc} 
\toprule

{\textbf{Attack}} & \textbf{Succ.} & \textbf{Pert.} & \textbf{Sim.}  & \textbf{Gram.}\\

\midrule
GA  & 50.9 & 5.0 & 79.3 & 0.9  \\
 PSO  & 60.3 & 3.7 & 81.8 & 0.7 \\
 \cdashline{1-5}[2pt/2pt]
\specialrule{0em}{1.5pt}{1.5pt}
HLBB  & 77.0 & 4.8 & 84.9 & 0.6 \\
TextHoaxer  & 78.8 & 5.1 & \textbf{85.8} & 0.6 \\
\textbf{\name} & \textbf{81.5} & \textbf{3.4} & 82.3 & \textbf{0.4}  \\
\bottomrule
\end{tabular}
\caption{Attack success rate (Succ., \%) $\uparrow$, perturbation rate (Pert., \%) $\downarrow$, average semantic similarity (Sim., \%) $\uparrow$, grammatical error increase rate (Gram., \%) $\downarrow$ of \name and the baselines on BERT using IMDB dataset under the query budget of 2,000.}
\vspace{-0.3cm}
\label{tb:semantic_performance}
\end{table}

\subsection{Evaluation on Adversary Quality}
\label{sec:adv_quality}

Adversarial examples should be indistinguishable from benign samples for humans but mislead the model prediction. Hence, textual adversarial examples should maintain the original meaning without apparent typos or grammatical errors. Though existing word-level attacks adopt synonym substitution to maintain semantic consistency, it is still possible to introduce grammatical error  and semantic inconsistency. Apart from the perturbation rate, we further evaluate the semantic similarity and grammatical error increase rate using the Universal Sequence Encoder (USE)~\cite{cer2018universal} and Language-Tool\footnote{\href{https://www.languagetool.org/}{https://www.languagetool.org/}}, respectively.

We compare \name with the baselines on BERT using IMDB dataset and summarize the results in Table~\ref{tb:semantic_performance}. With the lowest perturbation rate, \name exhibits better semantic similarity than the score-based attacks of GA and PSO but is lower than HLBB and TextHoaxer, which consider the semantic similarity of synonyms using the USE tool during the attack. However, USE tool is time-consuming and computationally expensive, resulting in HLBB and TextHoaxer running slower than \name as shown in Table~\ref{tb:real_world_application}, and their CPU occupancy rate is seven times that of \name. Also, \name achieves the lowest grammatical error  increase rate compared with the baselines. The human evaluation in Appendix~\ref{sec:human_evaluation} shows that the adversarial examples generated by \name are of high quality and difficult to be detected by humans. These evaluations demonstrate the high lexicality, semantic similarity and fluency of the generated adversarial examples of \name.

\begin{figure*}[t]
\centering
\includegraphics[width=\textwidth]{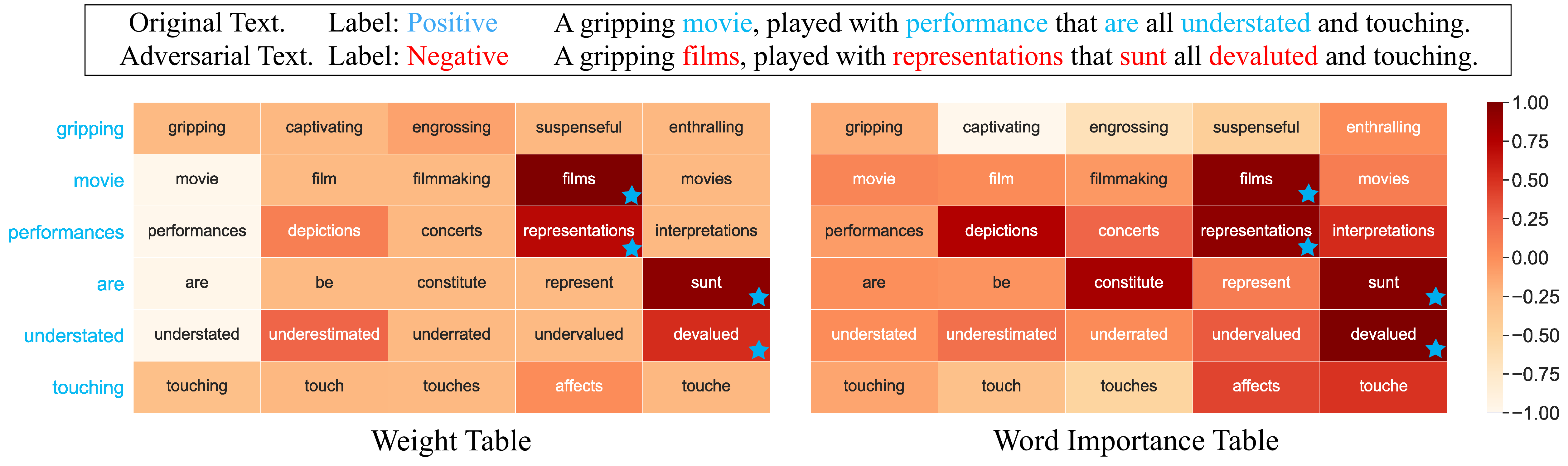}
\caption{Visualization of the weight table in \name and the word importance table from the victim model, representing the word importance of nouns, verbs, adjectives, adverbs, and their candidate words in the original text. The original words are highlighted in \origWord{Cyan}, with each row representing the candidate words. The substituted words are highlighted in \Red{Red} with marker \origWord{$\filledstar$}. 
A darker color  indicates a more important word. 
}
\label{fig:attention_table}
\end{figure*}

\subsection{Evaluation on Real-world Applications}

With the rapid development and broad application of DNNs, numerous companies have deployed many commercial Application Programming Interfaces (APIs) for various tasks, \eg sentiment analysis, named entity recognition, \etc. The user can  obtain the prediction label by calling the service API, making  it possible for hard-label attackers to attack.  To validate the attack effectiveness of \name in the real world, we evaluate the attack performance of \name, HLBB, and TextHoaxer on Amazon Cloud sentiment analysis API\footnote{\href{https://aws.amazon.com/}{https://aws.amazon.com/}}. Besides, attacks that run faster in the real world are more available and convenient. So we also report the average running time per attack.  Due to the high cost of commercial APIs, we sample 20 texts from IMDB dataset for the test. As shown in Table~\ref{tb:real_world_application}, \name achieves higher attack success rate, generates higher quality adversarial examples and runs faster than HLBB and TextHoaxer when facing real world APIs under tight query budget.

\begin{table}[tb]
\centering
\small
\setlength\tabcolsep{4pt}
\begin{tabular}{lccccc} 
\toprule

{\textbf{Attack}} & \textbf{Succ.} & \textbf{Pert.} & \textbf{Sim.}  & \textbf{Gram.} & \textbf{Time} \\

\midrule
HLBB  &  65.0 &  5.7 &  82.1 &  0.5 & 8.7 \\
TextHoaxer  &  65.0 &  5.2 &  \textbf{82.2} &  0.4 & 9.3 \\
\textbf{\name} &  \textbf{75.0} &  \textbf{3.1} &  80.9& \textbf{0.3} & \textbf{5.7}\\
\bottomrule
\end{tabular}
\caption{Attack success rate (Succ., \%) $\uparrow$, perturbation rate (Pert., \%) $\downarrow$, average semantic similarity (Sim., \%) $\uparrow$, grammatical error increase rate (Gram., \%) $\downarrow$, and running time per attack (Time, in minutes) $\downarrow$ of various hard-label attacks on Amazon Cloud APIs under the query budget of 2,000.}
\label{tb:real_world_application}
\vspace{-0.3cm}
\end{table}

\subsection{Visualization of Weight Table}
\label{sec:visualization}

Existing attacks~\cite{pwws,textfooler} usually take the model's output changes to different words as the word importance and perturb the top important words to generate adversarial examples. In this work, the weight table plays such a role, which learns the word importance from the attack history. Thus, the precise estimation of model's behavior is the key to generating better adversarial examples. To further explore \name, we conduct comparison and visualization to analyze the difference between the weight table and the word importance table from the model.
We generate the adversarial example of one benign text sampled from MR dataset by \name. For the word importance table, we calculate the word importance of each word by the prediction confidence difference after replacing the original word with the candidate word on BERT. We map the values in the learned weight table and word importance table into [-1, 1] and illustrate their heatmaps in  Figure~\ref{fig:attention_table}. More case studies are presented in Appendix~\ref{sec:case_study}.
We find that the weight table is consistent with the word importance table for the most important words. 
It helps \name optimize the adversarial perturbation more efficiently and hold on the most important words for better adversarial example.
This is important and challenging in the hard-label attack setting, which also explains the superiority of \name.

\subsection{Ablation Study}
To study the impact of different components of \name, we conduct a series of ablation studies on BERT using IMDB dataset under the query budget of 2,000.

\begin{table}[tb]
\centering
\small
\setlength\tabcolsep{4pt}
\begin{tabular}{lcccc} 
\toprule

{\textbf{Attack}} & \textbf{Succ.} & \textbf{Pert.} & \textbf{Sim.}  & \textbf{Gram.} \\

\midrule
Weight table  &  22.4 &  11.9 &  71.5 &  1.3 \\
Hybrid local search  &  79.6 &  ~~6.2 &  77.5 &  0.7 \\
\textbf{\name} &  \textbf{81.5} &  ~~\textbf{3.4} &  \textbf{82.3}& \textbf{0.4}\\
\bottomrule
\end{tabular}
\caption{Ablation study on the hybrid local search algorithm and weight table in \name  on BERT using IMDB dataset  under the query budget of 2,000.}
\label{tb:ablation_component_of_texthacker}
\vspace{-0.3cm}
\end{table}
\textbf{The impact of weight table and hybrid local search.} We design two variants to evaluate the impact of various components in \name. a) weight table: we remove the hybrid local search and greedily substitute the sampled word with its synonyms iteratively based on the weight table. b) Hybrid local search: we utilize the hybrid local search to search for better adversaries without weight table. The experiments in Table~\ref{tb:ablation_component_of_texthacker} show the effectiveness and rationality of different components in \name.

\begin{table}[tb]
\centering
\small
\setlength\tabcolsep{3pt}
\begin{tabular}{lcccc} 
\toprule

{\textbf{Attack}} & \textbf{Succ.} & \textbf{Pert.} & \textbf{Sim.}  & \textbf{Gram.} \\

\midrule
Local search $\to$ Mutation   &  79.1 &  6.1 &  77.5 &  0.7 \\
Recombination  $\to$ Crossover  &  81.3 &  3.7 &  81.9 &  0.4 \\
\textbf{\name} &  \textbf{81.5} &  \textbf{3.4} &  \textbf{82.3}& \textbf{0.4}\\
\bottomrule
\end{tabular}
\caption{Ablation study on the hybrid local search in \name and genetic algorithm in HLBB  on BERT using IMDB dataset  under the query budget of 2,000.}
\label{tb:ablation_hybrid_vs_genetic}
\vspace{-0.3cm}
\end{table}
\textbf{Hybrid local search vs. genetic algorithms.} Genetic algorithm in HLBB is inefficient in exploring the search space compared to the hybrid local search algorithm in \name that balances the local and global exploitation. Compared with random synonym substitutions on mutation in HLBB, the local search replaces more critical words using  word importance, making it reach the local optima faster. To further illustrate their differences, we replace local search with mutation and recombination with crossover respectively. The experiments in Table~\ref{tb:ablation_hybrid_vs_genetic} demonstrate that the first change drops the success rate by 2.4\% and increases the perturbation rate by 2.7\%. The second change drops the success rate by 0.2\% and increases the perturbation rate by 0.3\%. This study validates the better performance of local search and recombination.

\begin{table}[tb]
\centering
\small
\setlength\tabcolsep{4pt}
\begin{tabular}{lcccc} 
\toprule

{\textbf{Attack}} & \textbf{Succ.} & \textbf{Pert.} & \textbf{Sim.}  & \textbf{Gram.} \\

\midrule
Random-search  &  80.2 &  5.3 &  77.8 &  0.7 \\
Random-flip  &  81.0 &  5.3 &  76.4 &  0.7 \\
\textbf{\name} &  \textbf{81.5} &  \textbf{3.4} &  \textbf{82.3}& \textbf{0.4}\\
\bottomrule
\end{tabular}
\caption{Ablation study on the hybrid local search in \name and alternative strategies  on BERT using IMDB dataset  under the query budget of 2,000.}
\label{tb:ablation_local_search}
\vspace{-0.3cm}
\end{table}
\textbf{Local search vs. alternative strategies.} We replace the local search with two alternative strategies, namely random-search that randomly substitutes the sampled word with its synonyms, and random-flip that directly substitutes the sampled word with the original word. The experiments in Table~\ref{tb:ablation_local_search} demonstrate that local search achieves better attack performance than random-search and random-flip, showing the superiority of the local search in \name.

\section{Conclusion}
In this work, we propose a new text hard-label attack called \name.
\name captures the words that have higher impact on the adversarial example via the changes on prediction label.
By incorporating the learned word importance into the search process of the hybrid local search, \name can reduce the adversarial perturbation between the adversarial example and benign text more efficiently to generate more natural adversarial examples.
Extensive evaluations for two typical NLP tasks, namely text classification and textual entailment, using various datasets and models demonstrate that \name achieves higher attack success rate and lower perturbation rate than existing hard-label attacks and generates higher-quality adversarial examples. We believe that \name could shed new light on more precise estimation of the word importance and inspire more researches on hard-label attacks.

\section*{Limitations}
As shown in Table~\ref{tb:semantic_performance}, adversarial examples generated by \name have a slightly lower semantic similarity than HLBB and TextHoaxer from the automatic metric perspective. However, the quality (\ie lexicality, semantic similarity and fluency) of adversarial examples depend not only on semantic similarity evaluation, but also on perturbation rate, grammatical error rate, human evaluation, \etc. 
In our experiments, the quality in Table~\ref{tb:semantic_performance} and human evaluation experiment in Appendix~\ref{sec:human_evaluation} have demonstrated the higher quality and the harder detection by humans of the adversarial example generated by our \name. In addition, the semantic similarity metric is usually measured by the USE tool which will lead to high computing resource occupancy and slow running speed of the attack algorithm, as described in Section~\ref{sec:adv_quality}.  However, a faster and less resource-intensive attack attack is usually more suitable and convenient in the real world. Considering semantic similarity alone may not be a good choice for generating high quality adversarial examples. Hence, this limitation is acceptable.

\section*{Acknowledgement}
This work is supported by National Natural Science Foundation (62076105) and International Cooperation Foundation of Hubei Province, China (2021EHB011).

\bibliography{anthology,custom}
\bibliographystyle{acl_natbib}

\appendix
\newpage
\begin{figure*}[t]

\begin{subfigure}{.32\textwidth}
    \centering
    \includegraphics[width=\linewidth]{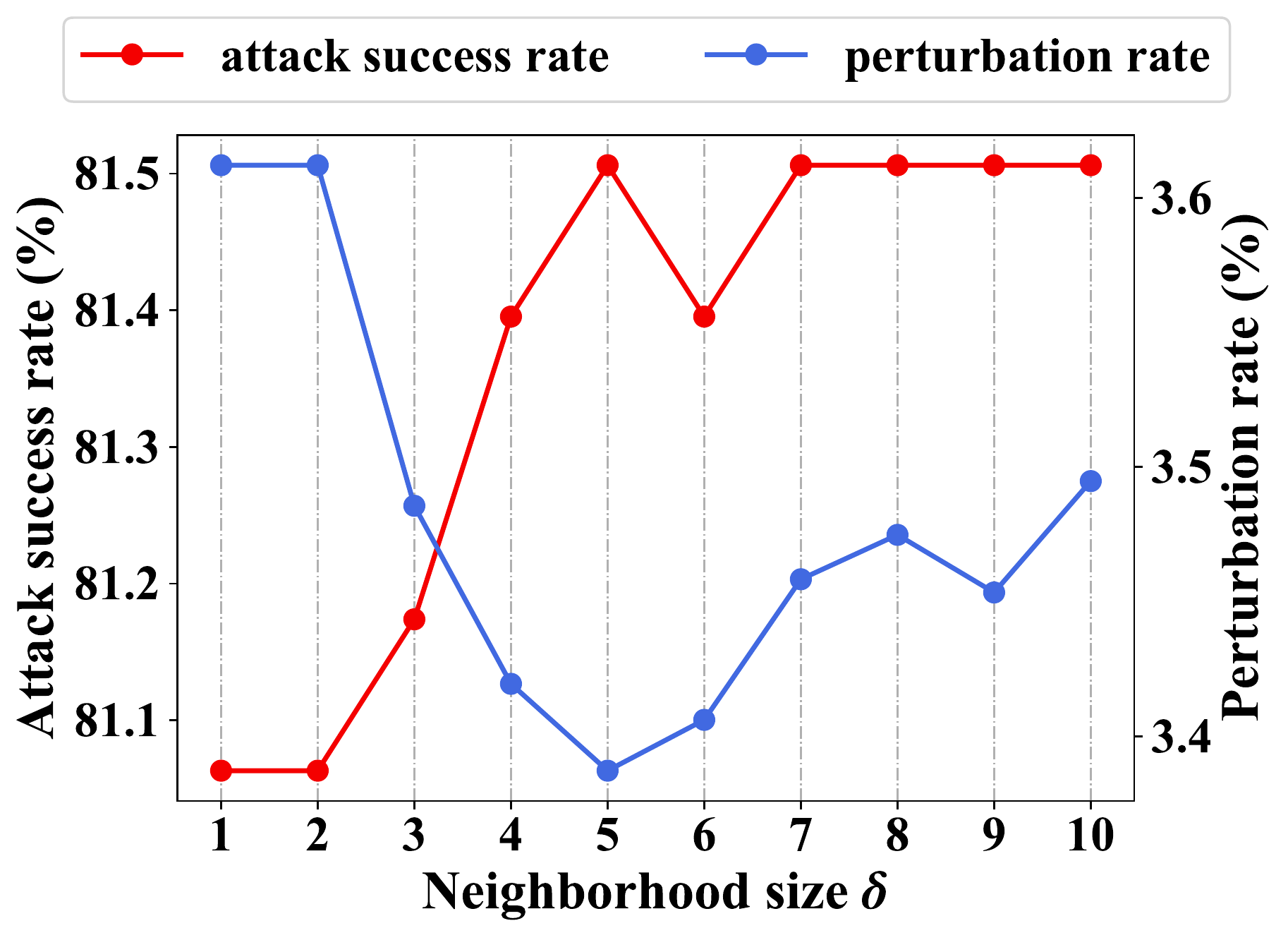}
    \caption{Parameter study for various $\delta$.}
    \label{fig:parameter:neighborhood}
\end{subfigure}%
\hspace{0.2cm}
\begin{subfigure}{.32\textwidth}
    \centering
    \includegraphics[width=\linewidth]{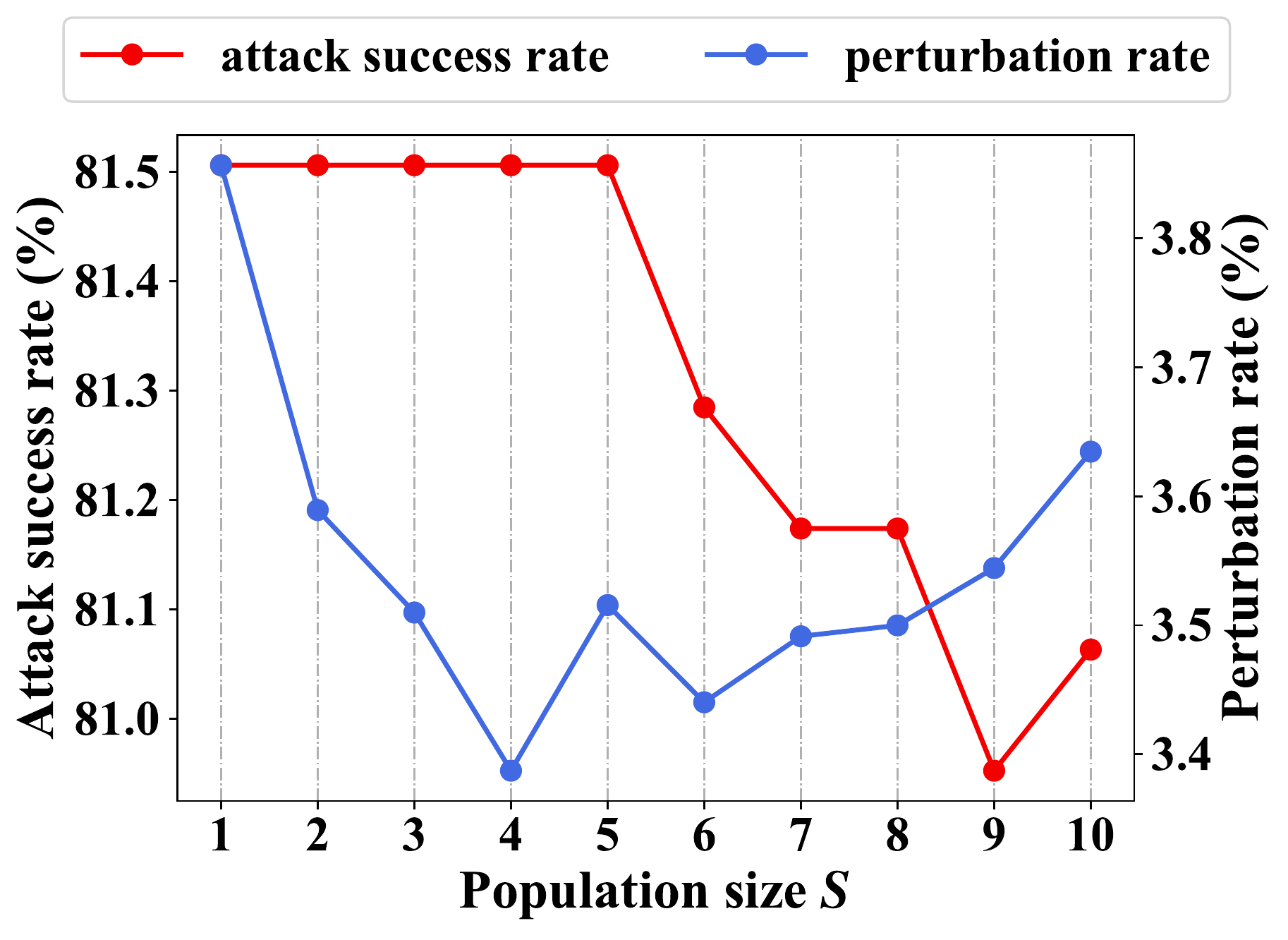}
    \caption{Parameter study for various $S$.}
    \label{fig:parameter:pop_size}
\end{subfigure}%
\hspace{0.2cm}
\begin{subfigure}{.32\textwidth}
    \centering
    \includegraphics[width=\linewidth]{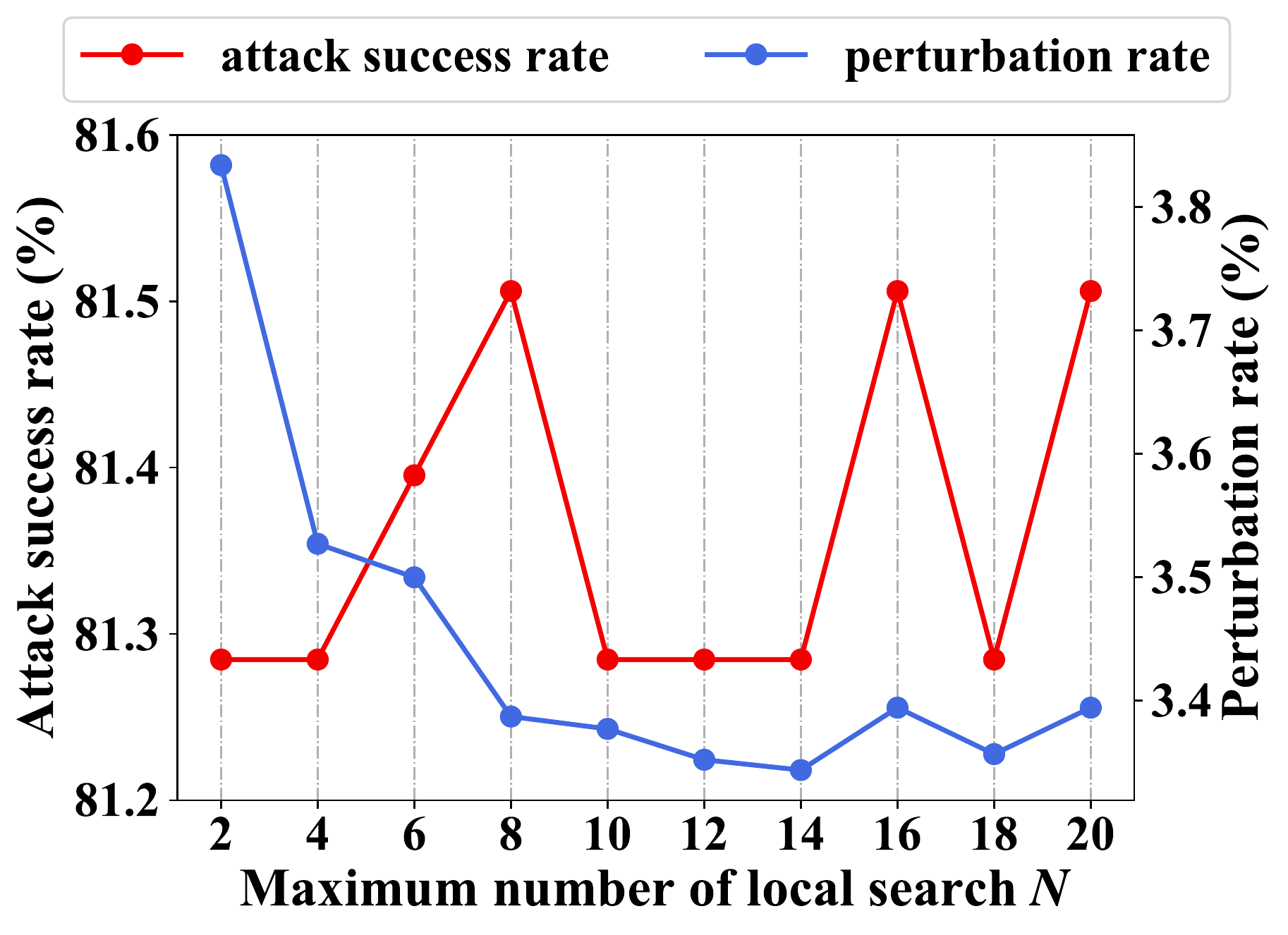}
    \caption{Parameter study for various $N$.}
    \label{fig:parameter:local_search}
\end{subfigure}%
\caption{The attack success rate (\%) $\uparrow$ and perturbation rate (\%) $\downarrow$ of \name on BERT using IMDB dataset, when varying the neighborhood size $\delta$, population size $S$ or maximum number of local search $N$.}
\label{fig:parameter_study}
\end{figure*}

\section{Parameter Study}
\label{appendix:parameter}
To gain more insights into the effectiveness of our \name, we conduct a series of parameter studies to explore the impact of hyper-parameters for the neighborhood size $\delta$, population size $S$, and maximum number of local search $N$ in \name. We conduct parameter studies on BERT using IMDB dataset to determine the best hyper-parameters and use the same hyper-parameters on all other datasets.

\textbf{On the neighborhood size.}
In Figure~\ref{fig:parameter:neighborhood}, we study the impact of the neighborhood size $\delta$. The small $\delta$ would restrict the search scope of local search, making it difficult to find the local optimal solution from the vast search space, resulting in low attack success rate and high perturbation rate under limited query budgets. As $\delta$ increases, the attack success rate increases and the perturbation rate decreases until $\delta = 5$. When we continually increase $\delta$, the vast search scope of local search makes it difficult to converge to local optima, resulting in an increase in perturbation rate. Thus, we set $\delta=5$ in our experiments.

\textbf{On the population size.}
As shown in Figure~\ref{fig:parameter:pop_size}, we study the impact of population size $S$. When $S=1$, the hybrid local search algorithm degrades to the non-population-based algorithm which exhibits high perturbation rate. With the increment on the value of $S$, the perturbation rate decreases until $S=4$. When we continually increase $S$, the local search operator costs many queries for each candidate solution in the population. This limits the number of iterations of the overall algorithm under tight query budget, leading to low attack success rate and high perturbation rate. Thus, we set $S=4$ in our experiments. 

\textbf{On the maximum number of local search.}
We finally study the impact of maximum number of local search $N$, as shown in Figure~\ref{fig:parameter:local_search}. When $N=2$, the recombination operator is performed for every two steps of the local search operator. It is difficult for local search operator to thoroughly explore the neighborhood space, resulting in low attack success rate and high perturbation rate. When $N$ is too large, there are too few recombination operations under tight budgets, making \name insufficient to explore the entire search space, leading to unstable performance. Therefore, we adopt an intermediate value $N=8$ to balance the local search and recombination in our experiments. 

\section{Why Do Population-based Baselines Perform Poor?}
\label{sec:baseline_analysis}
\begin{table}[tb]
\centering
\small
\begin{tabular}{lccccc} 
\toprule
\multirow{2.4}*{\textbf{Attack}} & \multicolumn{2}{c}{\footnotesize{\textbf{$S=4$}}} &&
\multicolumn{2}{c}{\footnotesize{\textbf{$S=30$}}} \\

\cmidrule(lr){2-3}\cmidrule(lr){5-6}

~   & \textbf{Succ.} & \textbf{Pert.}     && \textbf{Succ.} & \textbf{Pert.}       \\ 

\midrule
GA  & 88.2 & 9.4 && 35.5 & 3.4 \\ 
PSO & 75.6 & 6.4  && 47.3  & 2.8   \\
\cdashline{1-6}[2pt/2pt]
\specialrule{0em}{1.5pt}{1.5pt}
HLBB  & 65.3 & 4.5  && 77.0 & 4.8  \\
\textbf{\name}  & \textbf{81.5} & \textbf{3.4}  && \textbf{80.6} & \textbf{4.7}\\

\bottomrule
\end{tabular}
\caption{Attack success rate (Succ., \%) $\uparrow$, perturbation rate (Pert., \%) $\downarrow$ of \name and the baselines on BERT using IMDB dataset under the query budget of 2,000 when the population size $S=4$ and $S=30$.}
\label{tb:analysis_of_pop}
\end{table}

To further analyze why the baselines perform poorly under tight budgets, we show the performance of our \name and the population-based baselines on BERT using IMDB dataset under the same population size $S=4$ and $S=30$ (commonly used in GA, PSO and HLBB). Note that TextHoaxer is a non-population-based algorithm and is not considered in this experiment. As shown in Table~\ref{tb:analysis_of_pop}, when $S=4$, the low population size makes it difficult to seriously explore the search space and find the optimal adversarial example for GA and PSO, resulting in high perturbation rate. When $S=30$, GA and PSO cost too many queries in each iteration. Thus, tight budget makes it difficult for them to fully explore the entire search space to find adversarial examples, resulting in low attack success rate. In contrast, the adversary initialization by random walks ensures high attack success rate of \name and HLBB even under tight budgets. And the word importance learned by attack history helps \name explore more efficiently and obtain lower perturbation rate.

\section{Human Evaluation}
\label{sec:human_evaluation}
Human beings are very sensitive and subjective to texts. Even minor synonym substitutions may change the feeling of people, resulting in different evaluations.
Therefore, human evaluation is also necessary to evaluate the quality of adversarial examples.
We perform the human evaluation on 20 benign texts and the corresponding adversarial examples generated by \name, HLBB and TextHoaxer on BERT using MR dataset. Note that the texts in the MR dataset are shorter, averaging only 20 words per sentence, making it easier for humans to detect the adversarial examples. We invite 20 volunteers to label the adversarial examples, \ie positive or negative, and score for the similarity between the benign sample and its adversarial example from 1 (very similar) to 5 (very different). The survey results show that 84.5\% of the adversarial examples on \name (\vs 79.0\% on HLBB and 81.5\% on TextHoaxer) are labeled the same as the original samples, and the average similarity score is 1.9 (\vs 2.4 on HLBB and 2.1 on TextHoaxer). It demonstrates that the adversarial examples generated by \name are of higher quality and harder to be detected by humans than that of HLBB and TextHoaxer.

\section{More Visualizations of Weight Table}
\label{sec:case_study}

\begin{figure*}[t]
\centering
\includegraphics[width=\textwidth]{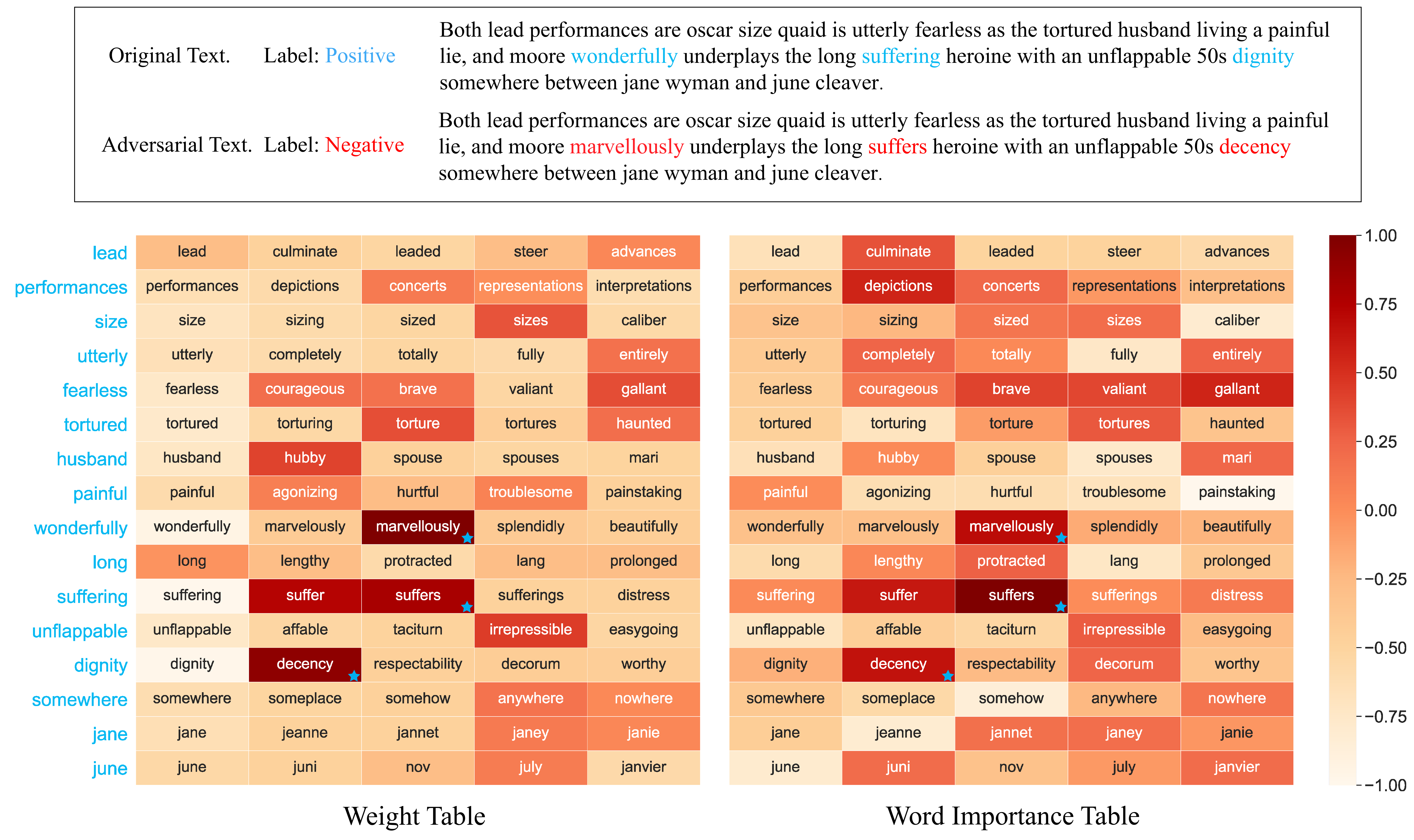}
\caption{Visualization of the weight table in \name and the word importance table from the victim model, representing the word importance of nouns, verbs, adjectives, adverbs, and their candidate words in the original text as shown in Table~\ref{tab:mr_sample}. The original words are highlighted in \origWord{Cyan}, with each row representing the candidate words. The substituted words are highlighted in \Red{Red} with marker \origWord{$\filledstar$}. 
A darker color indicates a more important word. 
}
\label{fig:mr_attention}
\end{figure*}

\begin{figure*}[t]
\centering
\includegraphics[width=\textwidth]{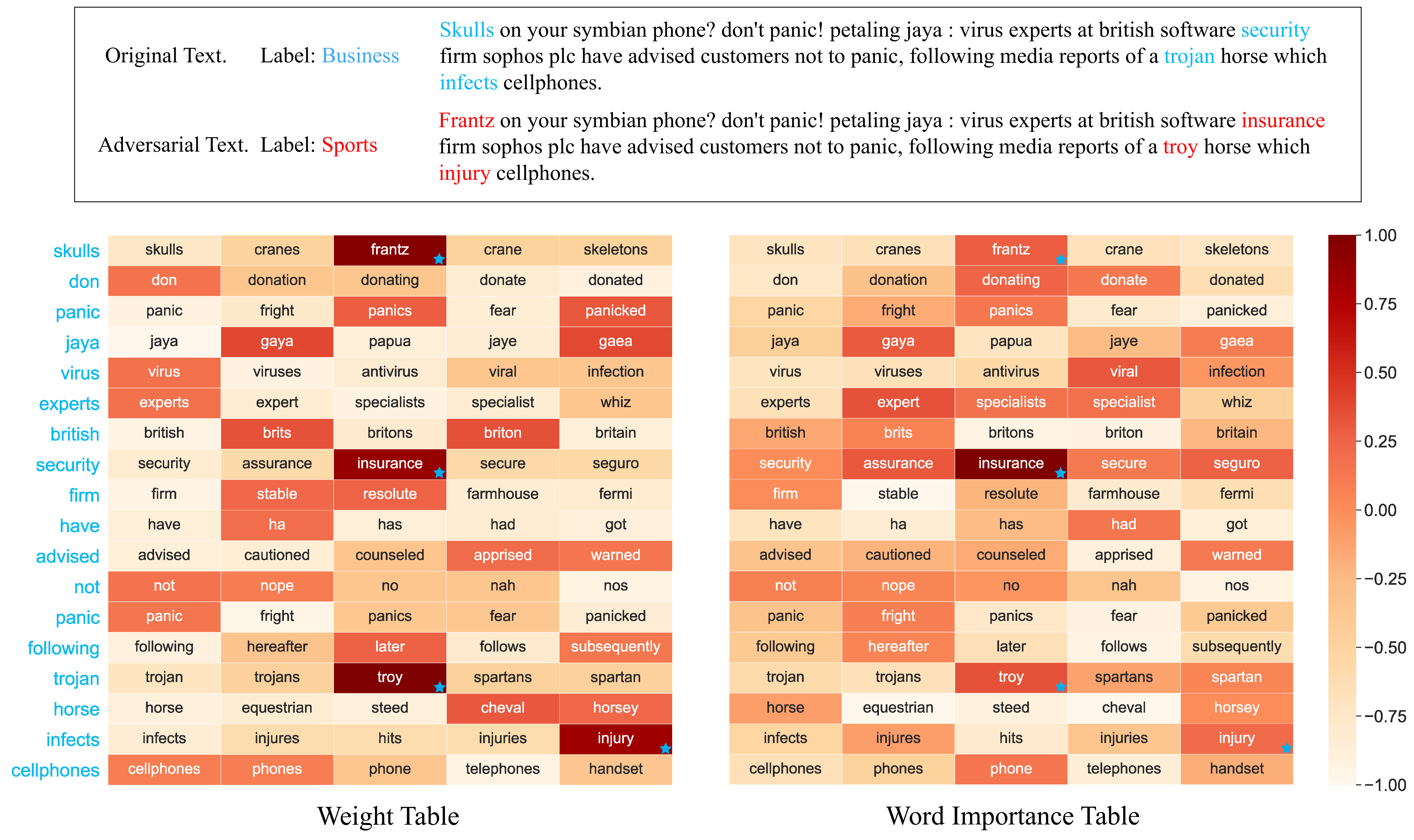}
\caption{Visualization of the weight table in \name and the word importance table from the victim model, representing the word importance of nouns, verbs, adjectives, adverbs, and their candidate words in the original text as shown in Table~\ref{tab:ag_sample}. The original words are highlighted in \origWord{Cyan}, with each row representing the candidate words. The substituted words are highlighted in \Red{Red} with marker \origWord{$\filledstar$}. 
A darker color indicates a more important word. 
}
\label{fig:ag_attention}
\end{figure*}

\begin{figure*}[t]
\centering
\includegraphics[width=\textwidth]{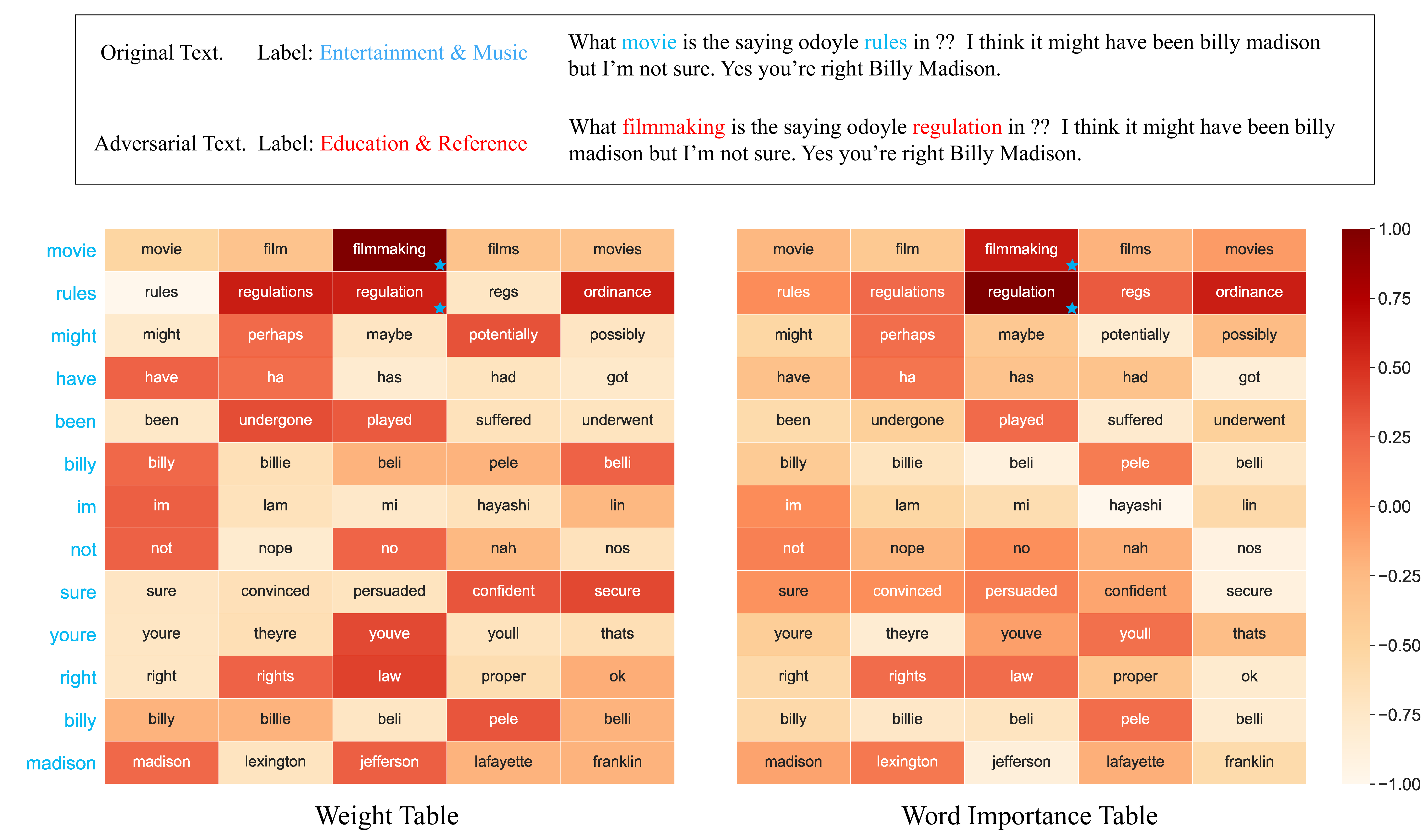}
\caption{Visualization of the weight table in \name and the word importance table from the victim model, representing the word importance of nouns, verbs, adjectives, adverbs, and their candidate words in the original text as shown in Table~\ref{tab:yahoo_sample}. The original words are highlighted in \origWord{Cyan}, with each row representing the candidate words. The substituted words are highlighted in \Red{Red} with marker \origWord{$\filledstar$}. 
A darker color indicates a more important word. 
}
\label{fig:yahoo_attention}
\end{figure*}

Here we present more case studies as the extension of Section~\ref{sec:visualization} in Figure~\ref{fig:mr_attention},~\ref{fig:ag_attention},~\ref{fig:yahoo_attention},  and the adversarial examples generated by various hard-label attacks in Table~\ref{tab:mr_sample},~\ref{tab:ag_sample}, ~\ref{tab:yahoo_sample}. These visualizations further verify the consistency between the weight table and the word importance table, proving the effectiveness of the learned weight table in \name.

\begin{table*}[tb]
    \small
    \centering
    \begin{tabular}{ccc}
    \toprule
    Attack & Original Text \& Adversarial Example & Prediction \\
    \midrule
    Original Text & \multicolumn{1}{m{11.6cm}}{Both lead performances are oscar size quaid is utterly fearless as the tortured husband living a painful lie, and moore wonderfully  underplays the long suffering heroine with an unflappable 50s dignity somewhere between jane wyman and june cleaver.}& \makecell{Positive} \\
    \specialrule{0em}{4pt}{4pt}
    
    HLBB  & \multicolumn{1}{m{11.6cm}}{Both \advAttack{lead}{leaded} performances are oscar size quaid is utterly \advAttack{fearless}{brave} as the \advAttack{tortured}{tortures} \advAttack{husband}{hubby} living a \advAttack{painful}{agonizing} lie, and moore wonderfully  underplays the long \advAttack{suffering}{suffer} \advAttack{heroine}{smack} with an unflappable 50s \advAttack{dignity}{decency} somewhere between jane wyman and june cleaver.}& \makecell{\Red{Negative}} \\
    \specialrule{0em}{4pt}{4pt}
    
    TextHoaxer  & \multicolumn{1}{m{11.6cm}}{Both lead performances are oscar size quaid is utterly fearless as the \advAttack{tortured}{tortures} \advAttack{husband}{hubby} living a \advAttack{painful}{agonizing} lie, and moore wonderfully  underplays the long \advAttack{suffering}{suffers} \advAttack{heroine}{smack} with an \advAttack{unflappable}{easygoing} 50s dignity \advAttack{somewhere}{nowhere} between jane wyman and june cleaver.}& \makecell{\Red{Negative}} \\
    \specialrule{0em}{4pt}{4pt}
    
    \name  & \multicolumn{1}{m{11.6cm}}{Both lead performances are oscar size quaid is utterly fearless as the tortured husband living a painful lie, and moore \advAttack{wonderfully}{marvellously}  underplays the long \advAttack{suffering}{suffers} heroine with an unflappable 50s \advAttack{dignity}{decency} somewhere between jane wyman and june cleaver.}& \makecell{\Red{Negative}} \\
    \bottomrule
    \end{tabular}
    \caption{The original text from MR dataset and the adversarial example generated by various hard-label attacks (HLBB, TextHoaxer and \name) on BERT. We highlight the words replaced by the attacks in \Red{Red}. The corresponding original words are highlighted in \origWord{Cyan}.}
    \label{tab:mr_sample}
\end{table*}

\begin{table*}[tb]
    \small
    \centering
    \begin{tabular}{ccc}
    \toprule
    Attack & Original Text \& Adversarial Example & Prediction \\
    \midrule
    Original Text & \multicolumn{1}{m{11.6cm}}{Skulls on your symbian phone? don't panic! petaling jaya : virus experts at british software security firm sophos plc have advised customers not to panic, following media reports of a trojan horse which infects cellphones.}& \makecell{Business} \\
    \specialrule{0em}{4pt}{4pt}
    
    HLBB  & \multicolumn{1}{m{11.6cm}}{Skulls on your symbian phone? don't panic! petaling jaya : \advAttack{virus}{infection} experts at british \advAttack{software}{sw} security firm sophos plc have advised customers not to panic, following media reports of a \advAttack{trojan}{spartans} horse which \advAttack{infects}{injury} \advAttack{cellphones}{telephones}.}& \makecell{\Red{Sports}} \\
    \specialrule{0em}{4pt}{4pt}
    
    TextHoaxer  & \multicolumn{1}{m{11.6cm}}{Skulls on your symbian phone? don't panic! petaling \advAttack{jaya}{gaya} : virus experts at british \advAttack{software}{sw} \advAttack{security}{insurance} \advAttack{firm}{resolute} sophos plc have advised customers not to panic, following media reports of a \advAttack{trojan}{spartans}  horse which infects cellphones.}& \makecell{\Red{Sports}} \\
    \specialrule{0em}{4pt}{4pt}
    
    \name  & \multicolumn{1}{m{11.6cm}}{\advAttack{Skulls}{Frantz} on your symbian phone? don't panic! petaling jaya : virus experts at british software \advAttack{security}{insurance} firm sophos plc have advised customers not to panic, following media reports of a \advAttack{trojan}{troy} horse which \advAttack{infects}{injury} cellphones.}& \makecell{\Red{Sports}} \\
    \bottomrule
    \end{tabular}
    \caption{The original text from AG’s News dataset and the adversarial example generated by various hard-label attacks (HLBB, TextHoaxer and \name) on BERT. We highlight the words replaced by the attacks in \Red{Red}. The corresponding original words are highlighted in \origWord{Cyan}.}
    \label{tab:ag_sample}
\end{table*}

\begin{table*}[tb]
    \small
    \centering
    \begin{tabular}{ccc}
    \toprule
    Attack & Original Text \& Adversarial Example & Prediction \\
    \midrule
    Original Text & \multicolumn{1}{m{10cm}}{What movie is the saying odoyle rules in ??  I think it might have been billy madison but I’m not sure. Yes you’re right Billy Madison.}& \makecell{Entertainment \& Music} \\
    \specialrule{0em}{4pt}{4pt}
    
    HLBB  & \multicolumn{1}{m{10cm}}{What \advAttack{movie}{filmmaking} is the \advAttack{saying}{proverb} odoyle rules in ??  I think it might have been billy madison but I’m \advAttack{not}{no} \advAttack{sure}{secure}. Yes you’re right Billy Madison.}& \makecell{\Red{Education \& Reference}} \\
    \specialrule{0em}{4pt}{4pt}
    
    TextHoaxer  & \multicolumn{1}{m{10cm}}{What \advAttack{movie}{filmmaking} is the \advAttack{saying}{proverb} odoyle rules in ??  I think it \advAttack{might}{perhaps} \advAttack{have}{ha} \advAttack{been}{undergone} billy madison but I’m not sure. Yes you’re right Billy Madison.}& \makecell{\Red{Education \& Reference}} \\
    \specialrule{0em}{4pt}{4pt}
    
    \name  & \multicolumn{1}{m{10cm}}{What \advAttack{movie}{filmmaking} is the saying odoyle \advAttack{rules}{regulation} in ??  I think it might have been billy madison but I’m not sure. Yes you’re right Billy Madison.}& \makecell{\Red{Education \& Reference}} \\
    \bottomrule
    \end{tabular}
    \caption{The original text from Yahoo! Answers dataset and the adversarial example generated by various hard-label attacks (HLBB, TextHoaxer and \name) on BERT. We highlight the words replaced by the attacks in \Red{Red}. The corresponding original words are highlighted in \origWord{Cyan}.}
    \label{tab:yahoo_sample}
\end{table*}

\label{appendix:case}

\end{document}